\documentclass[runningheads]{llncs}
\usepackage{graphicx}
\usepackage{cite}
\usepackage{amsmath,amssymb,amsfonts}
\usepackage{algorithmic}
\usepackage{graphicx}
\usepackage{textcomp}
\usepackage{xcolor}
\usepackage{enumerate}
\usepackage{graphicx}
\usepackage{url}
\usepackage{times}
\usepackage{latexsym}
\usepackage{amsmath,amssymb,amsfonts}
\usepackage{enumitem}
\usepackage{textcomp}
\usepackage{xcolor}
\usepackage{algorithm,algorithmic}
\usepackage{multirow}
\usepackage{booktabs}
\usepackage{array}
\usepackage{stfloats}
\usepackage{subfigure}
\usepackage{color}
\usepackage{caption2}
\usepackage{multicol}
\usepackage{cuted}
\usepackage[colorlinks]{hyperref}

\usepackage{amssymb}% http://ctan.org/pkg/amssymb
\usepackage{pifont}% http://ctan.org/pkg/pifont
\newcommand{\cmark}{\ding{51}}%
\newcommand{\xmark}{\ding{55}}%

\begin{document}
\title{Cross-domain error minimization for unsupervised domain adaptation}

\makeatletter
\newcommand{\printfnsymbol}[1]{%
  \textsuperscript{\@fnsymbol{#1}}%
}
\makeatother

\author{Yuntao Du \and
Yinghao Chen\thanks{equal contribution} \and
Fengli Cui \printfnsymbol{1}\and \\
Xiaowen Zhang \and
Chongjun Wang  \thanks{corresponding author}
}
\authorrunning{Yuntao Du et al.}
\institute{State Key Laboratory for Novel Software Technology at Nanjing University, \\ Nanjing University,  Nanjing 210023, China, \\ \email{duyuntao@smail.nju.edu.cn,yinghaochen48@gmail.com, \\ cuifengli1997@gmail.com, zhangxw@smail.nju.edu.cn, chjwang@nju.edu.cn}}

\maketitle              % typeset the header of the contribution
%

% both the feature distributions and labeling functions should be close across domains

\begin{abstract}
	Unsupervised domain adaptation aims to transfer knowledge from a labeled source domain to an unlabeled target domain. Previous methods focus on learning domain-invariant features to decrease the discrepancy between the feature distributions as well as minimizing the source error and have made remarkable progress. However, a recently proposed theory reveals that such a strategy is not sufficient for a successful domain adaptation. It shows that besides a small source error, both the discrepancy between the feature distributions and the discrepancy between the labeling functions should be small across domains. The discrepancy between the labeling functions is essentially the \textbf{cross-domain errors} which are ignored by existing methods. To overcome this issue, in this paper, a novel method is proposed to integrate all the objectives into a unified optimization framework. Moreover, the incorrect pseudo labels widely used in previous methods can lead to error accumulation during learning. To alleviate this problem,  the pseudo labels are obtained by utilizing structural information of the target domain  besides source classifier and we propose a curriculum learning based strategy to select the target samples with more accurate pseudo-labels during training. Comprehensive experiments are conducted, and the results validate that our approach outperforms state-of-the-art methods.

\end{abstract}
\vspace{-0.6cm}
\section{Introduction}
\vspace{-0.1cm}
Traditional machine learning methods have achieved significant progress in various application scenarios \cite{b3,b5}. Training a model usually requires a large amount of labeled data. However, it is difficult to collect annotated data in some scenarios, such as medical image recognition \cite{b6} and automatic driving \cite{b7}. Such a case may lead to performance degradation for traditional machine learning methods. \emph{Unsupervised domain adaptation} aims to overcome such challenge by transferring knowledge from a different but related domain (source domain) with labeled samples to a target domain with unlabeled samples \cite{b1}. And unsupervised domain adaptation based methods have achieved remarkable progress in many fields, such as image classification \cite{b32},  automatic driving \cite{b7} and medical image precessing \cite{b6}.

According to a classical theory of domain adaptation \cite{b2}, the error of a hypothesis $h$ in the target domain $\varepsilon_t(h)$ is bounded by three terms: the empirical error in the source domain $\hat{\varepsilon}_s(h)$, the distribution discrepancy across domains $d(\mathcal{D}_s, \mathcal{D}_t)$ and the ideal joint error  $\lambda^*$:
\begin{equation}
\varepsilon_t(h) \leq \hat{\varepsilon}_s(h) + d(\mathcal{D}_s, \mathcal{D}_t) + \lambda^*
\label{bound_ori}
\end{equation}
Note that $\mathcal{D}_s, \mathcal{D}_t$ denotes the source domain and the target domain, respectively. $\lambda^* = \varepsilon_s(h^*) +\varepsilon_t(h^*)$ is the ideal joint error and $h^* := \mathop{\arg\min}_{h \in \mathcal{H}} \varepsilon_s(h) + \varepsilon_t(h)$ is the ideal joint hypothesis. It is usually assumed that there is an ideal joint hypothesis $h^*$ which can achieve good performance in both domains, making $\lambda^*$ becoming a small and constant term. Therefore, besides minimizing the source empirical error, many methods focus on learning domain-invariant representations, i.e., intermediate features whose distributions are similar in the source and the target domain to achieve a small target error \cite{b13,b29,b34,b38,b54,b55,self1}. In shallow domain adaptation, \emph{distribution alignment} is a widely used strategy for domain adaptation \cite{b8,b10,b22,b23,b25}. These methods assume that there exists a common space where the distributions of two domains are similar and they concentrate on finding a feature transformation matrix that projects the features of two domains into a common subspace with less distribution discrepancy. 

Although having achieved remarkable progress, recent researches show that transforming the feature representations to be domain-invariant may inevitably distort the original feature distributions and enlarge the error of the ideal joint hypothesis \cite{b11,b12}. It reminds us that the error of the ideal joint error $\lambda^*$ can not be ignored. However, it is usually intractable to compute the ideal joint error $\lambda^*$, because there are no labeled data in the target domain. Recently, a general and interpretable generalization upper bound without the pessimistic term $\lambda^*$ for domain adaptation has been proposed in \cite{b14}:
\begin{equation}
\begin{aligned}
	\vspace{-0.1cm}
	\varepsilon_t(h) \leq \hat{\varepsilon}_s(h) + d(\mathcal{D}_s, \mathcal{D}_t) + \min \{E_{\mathcal{D}_s}[|f_s -f_t|],E_{\mathcal{D}_t}[|f_s -f_t|]\}
	\vspace{-0.4cm}
	\label{bound_new}
\end{aligned}
\end{equation}
where $f_s$ and $f_t$ are the labeling functions (i.e., the classifiers to be learned) in both domains. The first two terms in Equ (\ref{bound_new}) are similar compared with Equ (\ref{bound_ori}), while the third term is different. The third term measures the discrepancy between the labeling functions from the source and the target domain. Obviously, $E_{\mathcal{D}_s}[|f_s -f_t|] = \varepsilon_s(f_t)$ and $E_{\mathcal{D}_t}[|f_s -f_t|] = \varepsilon_t(f_s)$. As a result, the discrepancy between the labeling functions is essentially the \textbf{cross-domain errors}. Specifically, the cross-domain errors are the classification error of the source classifier in the target domain and the classification error of the target classifier in the source domain. Altogether, the newly proposed theory provides a sufficient condition for the success of domain adaptation: besides a small source error, not only the discrepancy between the feature distributions but also the cross-domain errors need to be small across domains, while the  cross-domain errors are ignored by existing methods.

Besides, estimating the classifier errors is important for domain adaptation. Various classifiers such as $k-$NN, linear classifier and SVMs have been used in shallow domain adaptation \cite{b8,b10,b15,b23}. Recently, some methods adopt the \emph{prototype classifier} \cite{b17} for classification in domain adaptation. The prototype classifier is a non-parametric classifier, where one class can be represented by one or more prototypes. And a sample can be classified according to the distances between the sample and the class prototypes.

% We aim at performing cross-domain error minimization to decrease the discrepancy between labeling functions. 
% complementary
In this paper, we propose a general framework named \emph{Cross-Domain Error Minimization} (CDEM) based on the prototype classifier. CDEM aims to simultaneously learn domain-invariant features and minimize the cross-domain errors,  besides minimizing the source classification error. To minimize the cross-domain errors, we maintain a classifier for each domain seperately, instead of assuming that there is an ideal joint classifier that can perform well in both domains. Moreover, we conduct discriminative feature learning for better classification. To sum up, as shown in Fig \ref{overall}, there are four objectives in the proposed method.   ($\romannumeral1$) Minimizing the classification errors in both domains to optimize the empirical errors. $(\romannumeral2$) Performing distribution alignment to decrease the discrepancy between feature distributions. $(\romannumeral3$) Minimizing the cross-domain errors to decrease the discrepancy between the labeling functions across domains. $(\romannumeral4$) Performing discriminative learning to learn discriminative features. Note that the objectives ($\romannumeral1$),  $(\romannumeral2$) and  $(\romannumeral4$) have been explored in previous methods \cite{b8,b22,b30}, while the objective ($\romannumeral3$) is ignored by existing methods. We integrate the four objectives into a unified optimization problem to learn a feature transformation matrix via a closed-form solution. After transformation, the discrepancy between the feature distributions and the cross-domain errors will be small, and the source classifier can generalize well in the target domain. 

Since the labels are unavailable in the target domain, we use \emph{pseudo labels} instead in the learning process. Inevitably, there are some incorrect pseudo labels, which will cause error accumulation during learning \cite{b55}. To alleviate this problem, the pseudo labels of the target samples are obtained based on  the structural information in the target domain and the source classifier, in this way, the pseudo labels are likey to be more accurate. Moreover, we propose to use \emph{curriculum learning} \cite{b19} based strategy to select target samples with high prediction confidence during training. We regard the samples with high prediction confidence as ”easy” samples and the samples with low prediction confidence as ”hard” samples.  The strategy is to learn the transformation matrix with "easy" samples at the early stage and with "hard" samples at the later stage.  With the iterations going on, we gradually add more and more target samples to the training process.

Note that CDEM is composed of two processes: learning transformation matrix and selecting target samples. We perform these two processes in an alternative manner for better adaptation. Comprehensive experiments are conducted on three real-world object datasets. The results show that CDEM outperforms the state-of-the-art adaptation methods on most of the tasks (16 out of 24), which validates the substantial effects of simultaneously learning domain-invariant features and minimizing cross-domain errors for domain adaptation.
\vspace{-0.55cm}
\section{Related Work}
\vspace{-0.35cm}
% Remarkable theoretical and algorithmic advances have been achieved in domain adaptation in recent years.

\textbf{Domain adaptation theory}. The theory in \cite{b2} is one of the pioneering theoretical works in this field. A new statistics named $\mathcal{H}\Delta \mathcal{H}$-divergence is proposed as a substitution of traditional distribution discrepancies (e.g. $L_1$ distance, KL-divergence) and a generalization error bound is presented. The theory shows that the target error is bounded by the source error and the distribution discrepancy across domains, so most domain adaptation methods aim to minimize the source error and reduce the distribution discrepancy across domains.  A general class of loss functions satisfying symmetry and subadditivity are considered in \cite{b64} and a new generalization theory with respect to the newly proposed discrepancy distance is developed. A margin-aware generalization bound based on asymmetric margin loss is proposed in \cite{b64} and reveals the trade-off between generalization error and the choice of margin. Recently, a theory considering labeling functions is proposed in \cite{b63}, which shows that the error of the target domain is bounded by three terms: the source error, the discrepancy in feature distributions and the discrepancy between the labeling functions across domains. The discrepancy between the labeling functions are essentially the cross-domain errors which are ingored by existing methods. CDEM is able to optimize all the objectives simultaneously.

\textbf{Domain adaptation algorithm}. The mostly used shallow domain adaptation approaches include instance reweighting \cite{b15,b20,b28} and distribution alignment \cite{b8,b10,b22,b30,b54}.

The instance reweighting methods assume that a certain portion of the data in the source domain can be reused for learning in the target domain and the samples in the source domain can be reweighted according to the relevance with the target domain. Tradaboost \cite{b20} is the most representative method which is inspired by Adaboost \cite{b27}. The source samples classified correctly by the target classifier have larger weight while the samples classified wrongly have less weight. LDML \cite{b28} also evaluates each sample and makes full use of the pivotal samples to filter out outliers. DMM \cite{b15} learns a transfer support vector machine via extracting domain-invariant feature representations and estimating unbiased instance weights to jointly minimize the distribution discrepancy. In fact, the strategy for selecting target samples based on \emph{curriculum learning} can be regarded as a special case of instance reweighting, where the weight of selected samples is 1, while the weight of unselected samples is 0.

The distribution alignment methods assume that there exists a common space where the distributions of two domains are similar and focus on finding a feature transformation that projects features of two domains into another latent shared subspace with less distribution discrepancy.  TCA \cite{b8} tries to align marginal distribution across domains, which learns a domain-invariant representation during feature mapping. Based on TCA, JDA \cite{b10} tries to align marginal distribution and conditional distribution simultaneously. Considering the balance between the marginal distribution and conditional distribution discrepancy, both BDA \cite{b30} and MEDA \cite{b22} adopt a balance factor to leverage the importance of different distributions. However, these methods all focus on learning domain-invariant features across domains and ignore the cross-domain errors. While our proposed method takes the cross-domain errors into consideration.

\vspace{-0.5cm}
\section{Motivation}
\vspace{-0.3cm}
\subsection{Problem Definition}
In this paper, we focus on unsupervised domain adaptation. There are a  source domain $\mathcal{D}_s = \{(x_s^i,y_s^i)\}_{i = 1}^{n_s}$ with $n_s$ labeled source examples  and a target domain $\mathcal{D}_t = \{{x_t^j}\}_{j= 1}^{n_t}$ with $n_t$ unlabeled target examples. It is assumed that the feature space and the label space are the same across domains, i.e., $\mathcal{X}_{s} = \mathcal{X}_{t} \in \mathbb{R}^d$, $\mathcal{Y}_s = \mathcal{Y}_t = \{1,2,...,C\}$, while the source examples and target examples are drawn from different joint distributions $P(\mathcal{X}_s,\mathcal{Y}_s)$ and $Q(\mathcal{X}_t,\mathcal{Y}_t)$, respectively. The goal of CDEM is to learn a feature transformation matrix $P \in R^{d \times k}$, which projects the features of both domains into a common space to reduce the shift in the joint distribution across domains, such that the target error $\varepsilon_{t}(h) = E_{(x,y) \sim Q}[h(x) \neq y]$ can be minimized, where $h$ is the classifier to be learned. 
\vspace{-0.3cm}
\subsection{Main Idea}
\vspace{-0.2cm}
% \setlength{\abovecaptionskip}{0.4cm}  % 调整图片标题与图距离
% \setlength{\belowcaptionskip}{-0.8cm}   % 调整图片标题与下文距离

\begin{figure*}[tbp]
	\begin{center}
	\includegraphics[width = 0.95\linewidth]{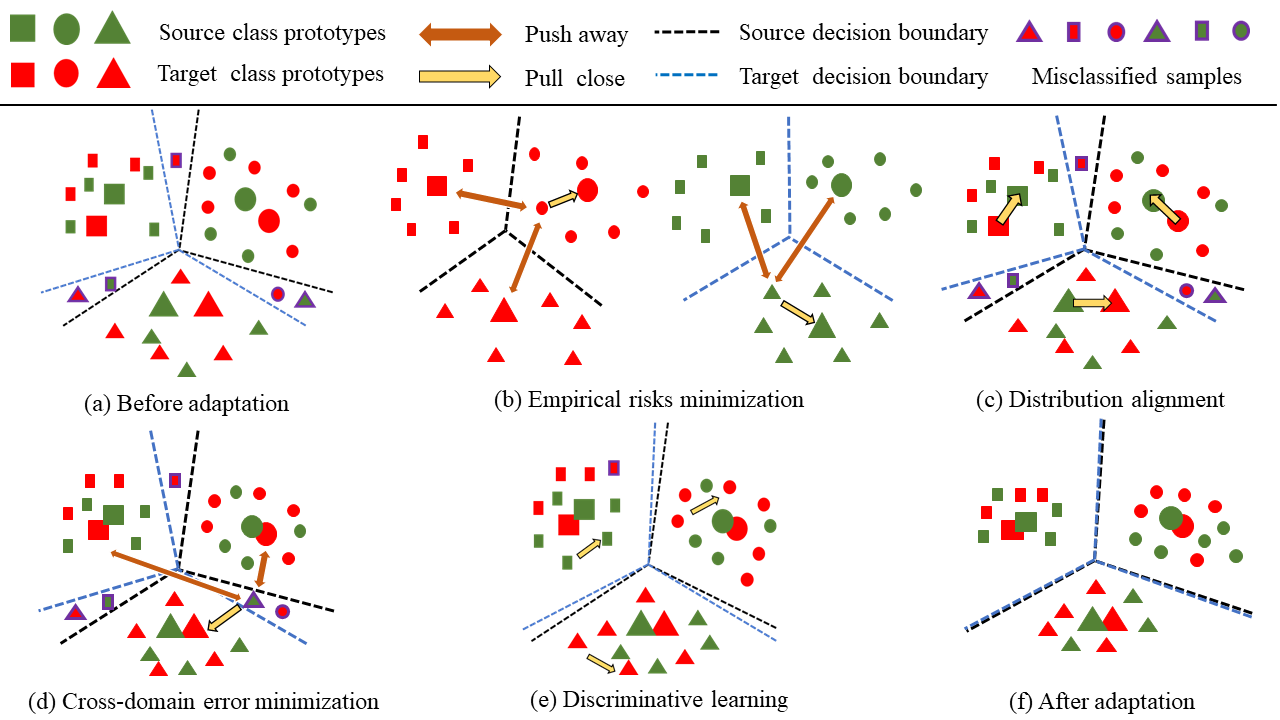}
	\caption{An overview of the proposed method. In this paper, we use the prototype classifier as the basic classifier. There is one prototype in each class, and we choose the class center as the prototype. We use the distances between samples and prototypes to calculate the classification error. (a) Before adaptation, the classifier trained in the source domain can not generalize well in the target domain. (b-e) We aim to minimize empirical errors in both domains, perform distribution alignment to learn domain-invariant features, minimize the cross-domain errors to pull the decision boundaries across domains close, and perform discriminative learning to learn discriminative features. (f) After adaptation, the discrepancy across domains is reduced, so that the target samples can be classified correctly by the source classifier.
    Best viewed in color.}
	\label{overall}
	\end{center}
\end{figure*}

% The cross-domain error is small when there is a small distribution discrepancy across domains, and the cross-domain error is large when there is a large discrepancy.
As shown in Fig \ref{overall} (a), there is a large discrepancy across domains before adaptation. Previous methods only focus on minimizing the source error and performing distribution alignment to reduce the domain discrepancy (Fig \ref{overall} (b-c)).
As the new theory revealed \cite{b14}, in addition to minimizing the source error and learning domain-invariant features, it is also important to minimize the cross-domain errors.  As shown in Fig \ref{overall}(d), although performing distribution alignment can reduce the domain discrepancy, the samples near the decision boundary are easy to be misclassified. Because performing distribution alignment only considers the discrepancy between the feature distributions, while the cross-domain errors are ignored. In the proposed method, minimizing the cross-domain errors can pull the decision boundaries  across domains close, so that we can obtain a further reduced domain discrepancy. Moreover, we also perform discriminative learning to learn discriminative features (Fig \ref{overall}(e)). Eventually, the domain discrepancy can be reduced and the classifier in the source domain can generalize well in the target domain (Fig \ref{overall}(f)). 

% complementary 
To sum up, we propose a general framework named \emph{cross-domain error minimization} (CDEM), which is composed of four objectives:
\begin{small}
\begin{equation}
	\vspace{-0.15cm}
	h = \arg\min\limits_{h \in \mathcal{H}} \sum_{i=1}^{n_s + n_t}l(h(x_i),y_i) + l_d(\mathcal{D}_s, \mathcal{D}_t) + l_f(\mathcal{D}_s, \mathcal{D}_t) + l_m(\mathcal{D}_s, \mathcal{D}_t)
	\vspace{-0.2cm}
\end{equation}
\end{small}
where $l(h(x_i),y_i) $ is the classification errors in both domains. $l_d(\mathcal{D}_s, \mathcal{D}_t)$ and $l_f(\mathcal{D}_s, \mathcal{D}_t)$ represent the discrepancy between the feature distributions and the discrepancy between the labeling functions across domains, respectively. $l_m(\mathcal{D}_s, \mathcal{D}_t)$ is the discriminative objective to learn discriminative features. Note that CDEM is a shallow domain adaptation method and use the prototype classifier as the classifier, where no extra parameters are learned except the transformation matrix $P$. The framework is general and can generalize to other methods such as deep models.

\setlength{\abovecaptionskip}{0.1cm}  % 调整图片标题与图距离
\setlength{\belowcaptionskip}{-0.9cm}   % 调整图片标题与下文距离

% \begin{enumerate}[label=(\arabic*)]
%     \item {minimizing the empirical error $\varepsilon_s(f_s)$ in the source domain and $\varepsilon_t(f_t)$ in the target domain}
%     \item {minimizing the distribution discrepancy $d(\mathcal{D}_s,\mathcal{D}_t)$ across domains including  the marginal distribution discrepancy $d_{m}(\mathcal{D}_s,\mathcal{D}_t)$ and the conditional distribution discrepancy $d_{c}(\mathcal{D}_s,\mathcal{D}_t)$} 
%     \item {minimizing the cross-domain errors $\varepsilon_s(f_t)$ in the source domain and $\varepsilon_t(f_s)$ in the target domain}
%     \item {minimizing the discrimnative loss $l_d$ in both domains}
% \end{enumerate}

As the labels in the target domain are unavailable, the \emph{pseudo labels} for the target data are used for training instead. However, they are always some incorrect pseudo labels 
% thus on the one hand making an unsatisfactory alignment, and on the other hand, 
and may lead to catastrophic error accumulation during learning. To alleviate this problem, we use the  \emph{curriculum learning} based strategy to select the target samples with more accurate pseudo labels which are obtained by taking advantage of source classifier and structural information of the target domain. With the iterations going on, we gradually add more and more target samples to the training process.
\vspace{-0.3cm}
\subsection{Classification Error}
\label{classif_error}
In this paper, we choose the prototype classifier \cite{b17} as the classifiers in both domains since the prototype classifier is a non-parametric classifier and is widely used in many tasks. As shown in Fig \ref{overall}, we maintain one prototype for each class and adopt prototype matching for classification. The class centers $\{\mu_c\}_{c=1}^C$ are used as the prototype of each class in this paper. And we denote the classifier in the source domain as $f_s$ and the classifier in the target domain as $f_t$. Given a training set $\mathcal{D} = \{x_i,y_i\}_{i=1}^{|{\mathcal{D}}|}$ with $|\mathcal{D}|$ samples, a sample $x \in \mathcal{D}$, the class center (prototype) for each class is defined as  $\mu_{c} = \frac{1}{n_c}\sum_{x_i \in \mathcal{D}_{c}}x_i$, where $\mathcal{D}_{c} = \{x_i:x_i \in \mathcal{D}, y(x_i) = c\}$ and $n_c = |\mathcal{D}_c|$. We can derive the conditional probability of a given sample $x$ belonging to class $y$ as:
\begin{small}
\begin{equation}
    p(y|x) = \frac{exp(-|| x - \mu_{y}||)}
    {\sum_{c=1}^{C} exp(-|| x - \mu_{c}||) }
\end{equation}
\end{small}
Assume the sample $x$ belongs to class $c$, it is expected that the conditional probability $p(y|x)$ is close to $[0,0,...,1,...,0]$, which is a $C$-dimensional one hot vector with the $c$-th dimension to be 1.  Our goal is to pull the sample close to the center of $c$-th class while push the sample away from other $C-1$ class centers. Note that instead of pushing samples directly away from $C-1$ centers, we view the data of other $C-1$ classes as a whole, and use the center of the $C-1$ classes $\widehat{\mu_{c}}$ to calculate the distance. As a result, the algorithm complexity can be reduced and the proposed algorithm can be accelerated. The objective of minimizing classification error can be represented as,
\begin{small}
\begin{equation}
	\min \sum_{(x,c) \sim \mathcal{D}} ||x -\mu_{c}||_{2}^{2} - \beta||x -\widehat{\mu_{c}}||^2_2
	\vspace{-0.3cm}
\end{equation}
\end{small}
where $\widehat{\mu_{c}} = \frac{1}{n_c^\star}\sum_{x_i \in \mathcal{D}/\mathcal{D}_{c}}x_i$  and $\widehat{\mu_{c}}$ is the center of all classes except class $c$ in the training set, $n_c^\star = |\mathcal{D}/\mathcal{D}_{c}|$, $\beta$ is the regularization parameter.
\vspace{-0.3cm}
\section{Method}
\vspace{-0.3cm}
% Our ultimate goal is to learn a feature transformation matrix $P$ to achieve these objectives simultaneously. 
In this section, we will describe all the objectives and the method to select target samples separately.
\vspace{-0.3cm}

\vspace{-0.2cm}
\subsection{Empirical Error Minimization}
For classifying the samples correctly, the first objective of CDEM is to minimize the empirical errors in both domains. Since there are no labeled data in the target domain, we use the pseudo labels \cite{b10} instead. The empirical errors in both domains are represented as,
\begin{footnotesize}
\begin{equation}
    \begin{aligned}
		\sum_{i=1}^{n_s + n_t}l(h(x_i),y_i) &= \varepsilon_s(f_s) +  \varepsilon_t(f_t) \\
		& = \sum_{c=1}^{C}\sum_{x_i\in{\mathcal{D}_{s,c}}}\left({||P^{T}(x_i-\mu_{s,c})||}_{2}^{2} - \beta||P^{T}(x_i-\widehat{\mu_{s,c}})||^2_2\right) \\
    & + \sum_{c=1}^{C}\sum_{x_j\in{\mathcal{D}_{t,c}}}\left({||P^{T}(x_j-\mu_{t,c})||}_{2}^{2} - \beta||P^{T}(x_j-\widehat{\mu_{t,c}})||^2_2\right)
	\end{aligned}
	\label{first_obj}
\vspace{-0.2cm}
\end{equation}
\end{footnotesize}
where $\mathcal{D}_{s,c} = \{x_i:x_i \in \mathcal{D}_s,  y(x_i) = c\}$ is the set of examples belonging to class $c$ in the source domain and $y(x_i)$ is the true label of $x_i$. Correspondingly,  $\mathcal{D}_{t,c} = \{x_j:x_j \in \mathcal{D}_t, \hat{y}(x_j) = c\}$ is the set of examples belonging to class $c$ in the target domain, where $\hat{y}(x_j)$ is the pseudo label of $x_j$. $\mu_{s,c} = \frac{1}{n_{s,c}}\sum_{x_i \in \mathcal{D}_{s,c}}x_i$ and $\mu_{t,c} = \frac{1}{n_{t,c}}\sum_{x_j \in \mathcal{D}_{t,c}}x_j$ are the centers of $c$-th class in the source domain and the target domain respectively, where $n_{s,c} = |\mathcal{D}_{s,c}|$ and $n_{t,c} = |\mathcal{D}_{t,c}|$. Similarly, $\widehat{\mu_{s,c}} = \frac{1}{n_{s,c}^\star}\sum_{x_i \in \mathcal{D}_s/\mathcal{D}_{s,c}}x_i$ and $\widehat{\mu_{t,c}} = \frac{1}{n_{t,c}^\star}\sum_{x_j \in \mathcal{D}_t/\mathcal{D}_{t,c}}x_j$ are the centers of all classes except class $c$ in the source domain and the target domain respectively, where $n_{s,c}^\star = |\mathcal{D}_s/\mathcal{D}_{s,c}|, {n_{t,c}^\star}= |\mathcal{D}_t/\mathcal{D}_{t,c}|$.

We further rewrite the first term of the objective function in Equ (\ref{first_obj}) as follows,
\begin{small}
\begin{equation}
	\begin{split}
	\vspace{-0.3cm}
	&\sum_{c=1}^{C}\sum_{x_i\in{\mathcal{D}_{s,c}}}\left({||P^{T}(x_i-\mu_{s,c})||}_{2}^{2}-\beta ||P^{T}(x_i-\widehat{\mu_{s,c})}||_{2}^{2} \right) \\
	& = \sum_{c=1}^{C}\left((1 - \beta)\sum_{x_i\in{\mathcal{D}_{s,c}}}{||P^{T}(x_i-\mu_{s,c})||}_{2}^{2} - \beta n_{s,c}   ||P^{T}(\mu_{s,c} -\widehat{\mu_{s,c})}||_{2}^{2} \right) \\
	&= (1 - \beta) \sum_{c=1}^{C}\sum_{x_i\in{\mathcal{D}_{s,c}}}{||P^{T}(x_i-\mu_{s,c})||}_{2}^{2} - \beta \sum_{c=1}^{C} n_{s,c}   {||P^{T}(\mu_{s,c} -\widehat{\mu_{s,c})}||_{2}^{2}} 
	\end{split}
	\label{first_obj_1}
	\vspace{-0.6cm}
\end{equation}
\end{small}

Inspired by Linear Discriminant Analysis (LDA) \cite{Mika1999FisherDA} and follow previous method \cite{Liang2019AggregatingRC}, we further transform the  two terms, which can be considered as intra-class variance in Equ (\ref{first_obj_1}), into similar expressions as Equ (\ref{first_obj_2}).
\begin{small}
\begin{equation}
	\vspace{-0.2cm}
	\begin{split}
&(1 - \beta)\sum_{c=1}^{C}\sum_{x_i\in{\mathcal{D}_{s,c}}}{||P^{T}(x_i-\mu_{s,c})||}_{2}^{2} -\beta \sum_{c=1}^{C} n_{s,c}||P^{T}(\mu_{s,c}-\widehat{\mu_{s,c}})||_{2}^{2} \\
&= \mathrm{tr}(P^{T}X_{s} (I - Y_{s}(Y_{s}^{T}Y_{s})^{-1} Y_{s}^{T}) X_{s}^{T}P) -\beta \sum_{c=1}^{C} n_{s,c}  \mathrm{tr}(P^{T}X_s \widehat{Q_{s,c}}X_s^{T}P)
	\end{split}
	\vspace{-0.2cm}
	\label{first_obj_2}
\end{equation}
\end{small}
Where $X_s \in \mathbb{R}^{d \times n_s}$ and $Y_s \in \mathbb{R}^{n_s \times C}$ are the samples and labels in the source domain. $tr(\cdot)$ is the trace of a matrix. By using target samples $X_t \in \mathbb{R}^{d \times n_s}$ and pseudo labels $\hat{Y}_t \in \mathbb{R}^{n_t \times C}$, the same strategy is also used to transform the second term in Equ (\ref{first_obj}). Denote $X = X_s \cup X_t \in \mathbb{R}^{d \times (n_s + n_t)}$, the objective of minimizing empirical errors can be written as,
\begin{footnotesize}
\begin{equation}
\begin{aligned}
\label{first_res}
\varepsilon_s(f_s) +  \varepsilon_t(f_t) = (1- \beta)(\mathrm{tr}(P^{T}XQ^{Y}X^TP) - \beta \sum_{c=1}^{C} \mathrm{tr}(P^{T}X \widehat{Q^{c}}X^{T}P))
\end{aligned}
\vspace{-0.2cm}
\end{equation}
\end{footnotesize}
where
\begin{footnotesize}
\begin{equation}
	\begin{split}
Q^{Y}= \begin{bmatrix}
    I - Y_{s} (Y_{s}^{T}Y_{s})^{-1} Y_{s}^{T} & \boldsymbol{0}\\ 
    \boldsymbol{0} & I - \hat{Y}_{t}(\hat{Y}_{t}^{T}\hat{Y}_{t})^{-1} \hat{Y}_{t}^{T}
\end{bmatrix}, 
\widehat{Q^{c}} = \begin{bmatrix}
    n_{s,c} \widehat{Q_{s,c}}& \boldsymbol{0}\\ 
    \boldsymbol{0} & n_{t,c} \widehat{Q_{t,c}}&
\end{bmatrix}
	\end{split}
\vspace{-0.6cm}
\end{equation}
\end{footnotesize}
\begin{small}
	\begin{equation}
	\begin{aligned}
	{(\widehat{Q_{s,c}})}_{ij} = \begin{cases}
		\frac{1}{n_{s,c}n_{s,c}}, & x_i,x_j \in \mathcal{D}_{s,c} \\
		\frac{1}{n_{s,c}^\star n_{s,c}^\star}, &x_i,x_j \in \mathcal{D}_s/\mathcal{D}_{s,c} \\
		-\frac{1}{n_{s,c}n_{s,c}^\star}, & otherwise
		\end{cases} , {(\widehat{Q_{t,c}})}_{ij} = \begin{cases}
			\frac{1}{n_{t,c}n_{t,c}}, & x_i,x_j \in \mathcal{D}_{t,c} \\
			\frac{1}{n_{t,c}^\star n_{t,c}^\star}, &x_i,x_j \in \mathcal{D}_t/\mathcal{D}_{t,c} \\
			-\frac{1}{n_{t,c}n_{t,c}^\star}, & otherwise
			\end{cases} 
	\end{aligned}
	\end{equation}
\end{small}

% \textbf{The detailed transforming process and the definition of the new variables $Q^{cc}$ and $\widehat{Q^{c}}$ are shown in the section Appendix-II.}
\vspace{-0.3cm}
\subsection{Distribution Alignment}
% \vspace{-0.2cm}
As there are feature distribution discrepancy across domains, the second objective of CDEM is to learn domain-invariant features for decreasing the discrepancy between feature distributions across domains. Distribution alignment is a popular method in domain adaptation \cite{b8,b10,b22}. To reduce the shift between feature distributions across domains, we follow \cite{b9} and adopt \emph{Maximum Mean Discrepancy} (MMD) as the distance measure to compute marginal distribution discrepancy $d_{m}(\mathcal{D}_s,\mathcal{D}_t)$ across domains  based on the distance between the sample means of two domains in the feature embeddings:
\begin{footnotesize}
	\begin{equation}
		\begin{aligned}
		d_{m}(\mathcal{D}_s,\mathcal{D}_t) = ||\frac{1}{n_s}\sum_{x_i \in \mathcal{D}_s}P^Tx_i - \frac{1}{n_t}\sum_{x_j \in \mathcal{D}_t}P^Tx_j||^2 = \mathrm{tr}(P^TXM_0X^TP)
		\end{aligned}
		\vspace{-0.2cm}
	\end{equation}
\end{footnotesize}
%, namely,
% \begin{small}
% \begin{equation}
%         MMD_{\mathcal{H}}^2[\mathcal{H},p,q] = ||E_p\phi(x) - E_q\phi(x)||_{\mathcal{H}}^2
% \end{equation}
% \end{small}
% Based on MMD, the marginal distribution discrepancy $d_{m}(\mathcal{D}_s,\mathcal{D}_t)$ is calculated as
% he conditional distribution discrepancy $d_c(\mathcal{D}_s,\mathcal{D}_t)$ is calculated as
Based on the pseudo labels of the target data, we minimize the conditional distribution discrepancy $d_c(\mathcal{D}_s,\mathcal{D}_t)$ between domains: 
\begin{small}
\begin{equation}
    \begin{aligned}
        d_{c}(\mathcal{D}_s,\mathcal{D}_t) = \sum_{c=1}^C||\frac{1}{n_{s,c}}\sum_{x_i \in \mathcal{D}_{s,c}}P^Tx_i - \frac{1}{n_{t,c}}\sum_{x_j \in \mathcal{D}_{t,c}}P^Tx_j||^2  = \sum_{c=1}^C\mathrm{tr}(P^TXM_cX^TP)
		\end{aligned}
		\vspace{-0.2cm}
\end{equation}
\end{small}
where,
\begin{small}
\begin{equation}
\begin{aligned}
{(\boldsymbol{M}_0)}_{ij} = \begin{cases}
    \frac{1}{n_s^2}, & x_i,x_j \in \mathcal{D}_s \\
    \frac{1}{n_t^2}, &x_i,x_j \in \mathcal{D}_t \\
    -\frac{1}{n_sn_t}, & otherwise
    \end{cases} , {(\boldsymbol{M}_c)}_{ij} = \begin{cases}
        \frac{1}{n_{s,c}^2}, & x_i,x_j \in \mathcal{D}_{s,c} \\
        \frac{1}{n_{t,c}^2}, &x_i,x_j \in \mathcal{D}_{t,c} \\
        -\frac{1}{n_{s,c}n_{t,c}}, & \begin{cases}
            x_i \in \mathcal{D}_{s,c}, x_j \in \mathcal{D}_{t,c} \\
            x_j \in \mathcal{D}_{s,c}, x_i \in \mathcal{D}_{t,c}
        \end{cases} \\
        0, & otherwise
    \end{cases}
\end{aligned}
\end{equation}
\end{small}
Denote $M = M_0 + \sum_{c=1}^CM_c$, then the objective of distribution alignment is equal to: 
\begin{small}
\begin{equation}
	 l_d(\mathcal{D}_s,\mathcal{D}_t) = d_{m}(\mathcal{D}_s,\mathcal{D}_t) + d_{c}(\mathcal{D}_s,\mathcal{D}_t) =  \mathrm{tr}(P^TXMX^TP) 
	 \vspace{-0.6cm}
\end{equation}
\end{small}
\vspace{-0.4cm}
\subsection{Cross-Domain Error Minimization}
% \vspace{-0.2cm}

Although performing distribution alignment can pull the two domains close, it is not enough for a good adaptation across domains. The discrepancy between the labeling functions, which is essentially the cross-domain errors, is another factor leading to the domain discrepancy \cite{b14} while is ignored by existing methods. Thus, the third objective of CDEM is to minimize cross-domain errors, by which the decision boundaries  across domains can be close and  the samples near the decision boundaries can be classified correctly, achieving a further reduced domain discrepancy and better adaptation.

% By which the the decision boundaries across domains can be more close and the samples near the decision boundaries can be classified correctly, achieving a further reduced domain discrepancy. 

% Similar to the first objective, we also aim at pulling the examples close to the corresponding class center and pushing them away from other $C-1$ class centers.

It is noticed that the cross-domain errors are the performance of the source classifier in the target domain and the performance of the target classifier in the source domain. As we use the prototype classifier, the cross-domain error in each domain is represented by the distances between the source samples (target samples) and the corresponding class centers in the target domain (source domain). For example, the cross-domain error in the source domain $\varepsilon_s(f_t)$ is the empirical error of applying the target classifier $f_t$ to the source domain $\mathcal{D}_s$.  Technically, the cross-domain errors in both domains are represented as,
\begin{footnotesize}
\begin{equation}
	\vspace{-0.2cm}
\label{third_ori}
    \begin{split}
	l_f(\mathcal{D}_s, \mathcal{D}_t) & = \varepsilon_s(f_t) + \varepsilon_t(f_s) \\
	& =  \sum_{c=1}^{C}\sum_{x_i\in{\mathcal{D}_{s,c}}}\left({||P^{T}(x_i-\mu_{t,c})||}_{2}^{2} - \beta||P^{T}(x_i-\widehat{\mu_{t,c}})||^2\right) 
    \\
    & + \sum_{c=1}^{C}\sum_{x_j\in{\mathcal{D}_{t,c}}}\left({||P^{T}(x_j-\mu_{s,c})||}_{2}^{2} - \beta||P^{T}(x_j-\widehat{\mu_{s,c}})||^2\right)
	\end{split} 
	\vspace{-0.4cm}
\end{equation}
\end{footnotesize}

Similar to the first objective, we transform the formula in Equ (\ref{third_ori}) as the following,
\begin{small}
	\begin{equation}
		\vspace{-0.2cm}
    \label{third_res}
        \begin{aligned}
\varepsilon_s(f_t) + \varepsilon_t(f_s) 
= &(1-\beta)\mathrm{tr}(P^{T}X Q^{Y}X^{T}P)+\sum_{c=1}^{C}n^{c}\mathrm{tr}(P^{T}XM_{c}X^{T}P)
\\
& - \beta\sum_{c=1}^{C}\mathrm{tr}(P^{T}X (n_{s,c}\widehat{Q_{s,t}^{c}}+n_{t,c}\widehat{Q_{t,s}^{c}})X^{T}P)
\end{aligned}
\vspace{-0.2cm}
\end{equation}
\end{small}
where 
\begin{small}
	\begin{equation}
	\begin{aligned}
	{(\widehat{Q_{s,t}^c})}_{ij} = \begin{cases}
		\frac{1}{n_{s,c}n_{s,c}}, & x_i,x_j \in \mathcal{D}_{s,c} \\
		\frac{1}{n_{t,c}^\star n_{t,c}^\star}, &x_i,x_j \in \mathcal{D}_{t,c} \\
		-\frac{1}{n_{s,c}n_{t,c}^\star},& \begin{cases}
			x_i \in \mathcal{D}_{s,c}, x_j \in \mathcal{D}_{t,c} \\
			x_j \in \mathcal{D}_{s,c}, x_i \in \mathcal{D}_{t,c}
		\end{cases} \\
		0 & otherwise
		\end{cases}
		{(\widehat{Q_{t,s}^c})}_{ij} = \begin{cases}
			\frac{1}{n_{t,c}n_{t,c}}, & x_i,x_j \in \mathcal{D}_{t,c} \\
			\frac{1}{n_{s,c}^\star n_{s,c}^\star}, &x_i,x_j \in \mathcal{D}_{s,c} \\
			-\frac{1}{n_{t,c}n_{s,c}^\star},& \begin{cases}
				x_i \in \mathcal{D}_{s,c}, x_j \in \mathcal{D}_{t,c} \\
				x_j \in \mathcal{D}_{s,c}, x_i \in \mathcal{D}_{t,c}
			\end{cases} \\
			0 & otherwise
			\end{cases}
	\end{aligned}
	\vspace{-0.6cm}
	\end{equation}
\end{small}

% \textbf{$M_c$ is defined in section IV.C. The detailed transforming process and the definition of the new variables $\widehat{Q^{c}_{s,t}}$ and $\widehat{Q^{c}_{t,s}}$ are shown in the section Appendix-II.}
\vspace{-0.2cm}
\subsection{Discriminative Feature Learning}
Learning domain-invariant features to reduce the domain discrepancy may harm the discriminability of the features \cite{b26}. So the fourth objective of CDEM is to perform discriminative learning to enhance the discriminability of the features \cite{b33}. To be specific, we resort to explore the structural information of all the samples to make the samples belonging to the same class close, which is useful for classification. Thus, the discriminative objective is,
\begin{small}
\begin{equation}
    l_m(\mathcal{D}_s, \mathcal{D}_t) = \sum_{x_i,x_j \in X}{||P^{T}x_{i}-P^{T}x_{j}||}_{2}^{2}{W}_{ij} 
\end{equation}
\end{small}
where $W \in \mathbb{R}^{(n_s+n_t) \times (n_s + n_t)}$ is the similarity matrix, which is defined as follows,
\begin{small}
\begin{equation}
\begin{aligned}
{W}_{ij} = \begin{cases}
    1, & y_i (\hat{y}_i) = y_j(\hat{y}_j) \\
    0, & y_i (\hat{y}_i) \neq y_j (\hat{y}_j) \\
    \end{cases}  
        \end{aligned}
	\end{equation}
\end{small}
This objective can be transformed as follows,
\begin{small}
\begin{equation}
    \begin{aligned}
    \sum_{x_i,x_j \in X}{||P^{T}x_{i}-P^{T}x_{j}||}_{2}^{2}{W}_{ij}
    &= \mathrm{tr}(\sum_{x_i \in X}P^{T}x_{i}{B}_{ii}x_{i}^{T}P - \sum_{x_i,x_j \in X}P^{T}x_{i}{W}_{ij}x_{j}^{T}P)\\
    &= \mathrm{tr}(P^{T}X B X^{T}P - P^{T}X W X^{T}P)
    = \mathrm{tr}(P^{T}X L X^{T}P)
\end{aligned}
\end{equation}
\end{small}
Where, $L = B - W \in \mathbb{R}^{(n_s + n_t) \times (n_s + n_t)}$ is the laplacian matrix,and $B \in \mathbb{R}^{(n_s + n_t) \times (n_s + n_t)}$ is a diagonal
matrix with ${(B)}_{ii} = \sum_{j}{(W)}_{ij}$.

\vspace{-0.2cm}
\subsection{Optimization}
Combining the four objectives together, we get the following optimization problem,
\vspace{-0.1cm}
\begin{small}
\begin{equation}
    \begin{aligned}
    L(p) & = \mathrm{tr}(P^{T}X Q^{Y} X^{T}P)- \beta \sum\limits_{c=1}^{C} \mathrm{tr}(P^{T}X \widehat{Q^{c}}X^{T}P) 
	+ \lambda \mathrm{tr}(P^{T}XMX^{T}P) \\
	&-\gamma \sum_{c=1}^{C} \mathrm{tr}(P^{T}X(\widehat{Q_{s,t}^{c}}+\widehat{Q_{t,s}^{c}})X^{T}P))
    +\eta \mathrm{tr}(P^{T}X L X^{T}P)+\delta||P||_{F}^{2}\\
    &=\mathrm{tr}(P^{T}X\Omega X^{T}P)+\delta||P||_{F}^{2}\\
    &s.t. \quad  P^{T}X H X^{T}P = I\\
    \end{aligned}
    \label{optim}
\end{equation}
\end{small}
where
% \begin{small}
%     \begin{equation}
%         \begin{aligned}
$\Omega =  Q^{Y} + \lambda M  +\eta L - \sum_{c=1}^{C}( \beta \widehat{Q^{c}}+ \gamma \widehat{Q_{s,t}^{c}} + \gamma \widehat{Q_{t,s}^{c}})$ and $H = \textbf{I} - \frac{1}{n_s+n_t}\textbf{1}$ is the centering matrix.
% \end{aligned}
% \end{equation}
% \end{small}
According to the constrained theory, we denote $\Theta = diag(\theta_1,...,\theta_k) \in R^{k \times k}$ as the Langrange multiplier, and derive the Langrange function for problem (\ref{optim}) as,
\begin{small}
    \begin{equation}
        \begin{aligned}
        L = \mathrm{tr}(P^{T}X\Omega X^{T}P)+\delta||P||_{F}^{2} + \mathrm{tr}((I -P^{T}X H X^{T}P)\Theta)\\
        \end{aligned}
        \label{optim_sol}
    \end{equation}
    \end{small}
Setting $\frac{\partial L}{\partial P} = \bold{0}$, we get generalized eigendecomposition,
\begin{small}
\begin{equation}
	(X\Omega X^T + \delta I) P = X HX^{T}P\Theta
\label{eig}
\end{equation}
\end{small}
Finally, finding the optimal feature transformation matrix $P$ is reduced to solving Equ (\ref{eig}) for the $k$ smallest eigenvectors.  
\vspace{-0.3cm}
\subsection{Selective target samples}
\vspace{-0.1cm}
% \textbf{Pseudo-labeling via Nearest Class Prototype (NCP)\cite{b35} and Target Clustering}
% We use the NCP and target domain clustering for pseudo-labeling. Firstly, we predict the pseudo label for the target sample using NCP, then utilize target clustering to obtain another pseudo label. Finally pseudo-labeling is conducted via combining these two predictions.
 To avoid the catastrophic error accumulation caused by the incorrect pseudo labels, we predict the pseudo labels for the target samples via exploring the structural information of the target domain and source classifier. Moreover, based on {curriculum learning}, we propose a strategy to select a part of target samples, whose pseudo labels are more likely to be correct, to participate in the next iteration for learning the transformation matrix. One simple way to predict pseudo labels for target samples is to  use the source class centers $\{\mu_c\}_{c=1}^C$ (the prototypes for each class) to classify the target samples. Therefore the conditional probability of a given target sample $x_t$ belonging to class $y$ is defined as:
\begin{small}
\begin{equation}
    p_{s}(y|x_{t}) = \frac{exp(-P^{T}|| x_{t} - \mu_{s,y}||)}
	{\sum_{c=1}^{C} exp(- P^{T}|| x_{t} - \mu_{s,c}||) }
	\vspace{-0.1cm}
\end{equation}
\end{small}

Because there exists distribution discrepancy across domains, only using source prototypes is not enough for pseudo-labeling, which will lead to some incorrect pseudo labels. We further consider the structural information in the target domain, which can be exploited by unsupervised clustering. In this paper, K-Means clustering is used in the target domain. The cluster center $\mu_{t,c}$ is initialized with corresponding class center $\mu_{s,c}$ in the source domain, which ensures one-to-one mapping for each class. Thus, based on target clustering, the conditional probability of a given target sample $x_t$ belonging to class $y$ is defined by:
\begin{small}
\begin{equation}
    p_{t}(y|x_{t}) = \frac{exp(- P^{T}|| x_{t} - \mu_{t,y}||)}
	{\sum_{c=1}^{C} exp(-P^{T}|| x_{t} - \mu_{t,c}||) }
	\vspace{-0.2cm}
\end{equation}
\end{small}

After getting $p_{s}(y|x_t)$ and $p_{t}(y|x_t)$,  we can obtain two different kinds of pseudo labels $\hat{y}_s^t$ and $\hat{y}_t^t$ for target samples $x_{t}$:
 \begin{small}
 \begin{equation}
	 \begin{aligned}
	 \hat{y}_{s}^{t} =\mathop{\arg\max}_{y \in {Y_t}} p_{s}(y|x_{t}) \quad
	 \hat{y}_{t}^{t} =\mathop{\arg\max}_{y \in {Y_t}} p_{t}(y|x_{t})
	 \end{aligned}
	 \vspace{-0.4cm}
 \end{equation}
 \end{small} 

 Based on these two kinds of pseudo labels, a curriculum learning based strategy is proposed to select a part of target samples for training. We firstly select the target samples whose pseudo labels predicted by $p_{s}(y|x_t)$ and $p_{t}(y|x_t)$ are the same (i.e., $\hat{y}_{s}^{t}= \hat{y}_{t}^{t}$). And these samples are considered to satisfy the \emph{label consistency} and are likely to be correct. Then, we progressively select a subset containing top $tn_t/T$ samples with highest prediction probabilities from the samples satisfying the label consistency, where $T$ is the number of total iterations and $t$ is the number of current iteration.  Finally, we combine $p_s(y|x_t)$ and $p_t(y|x_t)$ in an iterative weighting method. Formally, the final class conditional probability and the pseudo label for $x_t$ are as follows:
\begin{small}
\begin{equation}
	\begin{aligned}
    p(y|x_{t}) = (1 - &t/T) \times p_{s}(y|x_{t}) + t/T \times p_{t}(y|x_{t}) \\
		& \hat{y}_{t} =\mathop{\arg\max}_{y \in {Y_t}} p(y|x_{t})
	\end{aligned}
	\vspace{-0.4cm}
\end{equation}
\end{small}

To avoid the class imbalance problem when selecting samples, we take the class-wise selection into consideration to ensure that each class will have a certain proportion of samples to be selected, namely, 
\begin{small}
\begin{equation}
	N_{t,c}= \min(n_{t,c} \times t/T, n_{t,c}^{con})
	\vspace{-0.2cm}
\end{equation}
\end{small}
where $N_{t,c}$ is the number of target samples being selected of class $c$, $n_{t,c}^{con}$ denotes the number of target samples satisfying the label consistency in the class $c$ and $t$ is the current epoch.

\textbf{Remark}: CDEM is composed of two processes: learning transformation matrix $P$ and selecting target samples. We firstly learn the transformation matrix $P$ via solving the optimization problem (\ref{eig}). Then, we select the target sampels in the transformed feature space. We perform the two processes in an alternative manner as previous method \cite{b55}. 
% \begin{algorithm}[tp]
% 	\caption{\textbf{CDEM} }
% 	\label{algo_1}
% 	\begin{algorithmic}[1]
% 	\REQUIRE source samples $X_s$, source label $Y_s$, target samples $X_t$ iterations number $T$, parameters $\beta, \lambda, \gamma,\eta, \delta$ and the dimension of PCA space $m$ and dimension of CDEM space $m$. 
% 	\ENSURE Feature transformation matrix $P$.
% 	\STATE Train a base classifier using $\mathcal{D}_s$ and then apply prediction on $\mathcal{D}_t$ to get its pseudo labels $\hat{y}_t$.
% 	\STATE Construct kernel matrix $K$, graph Laplacian matrix $L$ by equation (\ref{cal_L}).
% 	\FOR{$t = 1,2,...,T$}
% 	   \STATE Calculate the balance factor $\mu$ using equation (\ref{cal_mu}) and construct MMCD matrix $V$ by equation (\ref{cal_M}),(\ref{cal_Z}).
% 	   \STATE Calculate $\beta_{init}$ by solving equation (\ref{cal_init}) and obtain adaptive classifier $f$ by equation (\ref{cal_F}),
% 	   \STATE Update the pseudo labels of $\mathcal{D}_t$ : $\hat{y}^t = f(X_t)$.
% 	\ENDFOR	
% 	\end{algorithmic}
% \end{algorithm}

\vspace{-0.4cm}
\section{Experiment}
\vspace{-0.2cm}
In this section, we evaluate the performance of CDEM by extensive experiments on three widely-used common datasets. The source code of CDEM is available at {\color{blue}\url{https://github.com/yuntaodu/CDEM}}.
\vspace{-0.5cm}
\subsection{Data Preparation}
\vspace{-0.2cm}
The \textbf{Office-Caltech} dataset\cite{b21} consists of images from 10 overlapping object classes between Office31 and Caltech-256 \cite{Griffin2007Caltech256OC}.
Specifically, we have four domains, \textbf{C} (\emph{Caltech-256}), \textbf{A} (\emph{Amazon}), \textbf{W} (\emph{Webcam}), and \textbf{D} (\emph{DSLR}).  By randomly selecting two different domains as the source domain and target
domain respectively, we construct $3 \times 4 = 12$ cross-domain object tasks, e.g. \textbf{C} $\rightarrow$ \textbf{A}, \textbf{C} $\rightarrow$ \textbf{W},..., \textbf{D} $\rightarrow$ \textbf{W}.

The \textbf{Office-31} dataset \cite{Saenko2010AdaptingVC} is a popular benchmark for visual domain adaptation. The dataset contains three real-world object domains, \emph{Amazon} (\textbf{A}, images downloaded from online merchants), \emph{Webcom} (\textbf{W}, low-resolution images by a web camera), and \emph{DSLR} (\textbf{D}, high-resolution images by a digital camera). It has 4652 images of 31 classes. We evaluate all methods on six transfer tasks: \textbf{A} $\rightarrow$ \textbf{W}, \textbf{A} $\rightarrow$ \textbf{D}, \textbf{W} $\rightarrow$ \textbf{A}, \textbf{W} $\rightarrow$ \textbf{D}, \textbf{D} $\rightarrow$ \textbf{A}, and \textbf{D} $\rightarrow$ \textbf{W}.

\textbf{ImageCLEF-DA}\footnote {{\color{blue}http://imageclef.org/2014/adaptation}} is a dataset organized by selecting 12 common classes shared by three public datasets, each is considered as a domain: \emph{Caltech-256} (\textbf{C}), \emph{ImageNet ILSVRC 2012} (\textbf{I}), and \emph{Pascal VOC 2012} (\textbf{P}). We evaluate all methods on six transfer tasks: \textbf{I} $\rightarrow$ \textbf{P}, \textbf{P} $\rightarrow$ \textbf{I}, \textbf{I} $\rightarrow$ \textbf{C}, \textbf{C} $\rightarrow$ \textbf{I}, \textbf{C} $\rightarrow$ \textbf{P}, and \textbf{P} $\rightarrow$ \textbf{C}.

\vspace{-0.5cm}
\subsection{Baseline Methods}
\vspace{-0.2cm}
We compare the performance of CDEM with several state-of-the-art traditional and deep domain adaptation methods:
\begin{itemize}
	\item Traditional domain adaptation methods: \textbf{1NN} \cite{Delany2007kNearestNC},\textbf{SVM} \cite{Evgeniou2001SupportVM} and \textbf{PCA} \cite{Hotelling1933AnalysisOA}, Transfer Component Analysis (\textbf{TCA}) \cite{b8}, Joint Distribution Alignment (\textbf{JDA}) \cite{b10}, CORrelation Alignment (\textbf{CORAL}) \cite{b54}, Joint Geometrical and Statistical Alignment (\textbf{JGSA}) \cite{b26}, Manifold Embedded Distribution Alignment (\textbf{MEDA}) \cite{b22}, Confidence-Aware Pseudo Label Selection (\textbf{CAPLS}) \cite{b31} and Selective Pseudo-Labeling (\textbf{SPL}) \cite{b55}.
	\item  Deep domain adaptation methods: Deep Domain Confusion (\textbf{DDC}) \cite{b29}, Deep Adaptation Network (\textbf{DAN}) \cite{b9}, Deep CORAL (\textbf{DCORAL}) \cite{b25}, Residual Transfer Network (\textbf{RTN}) \cite{b40}, Multi Adversarial Domain Adaptation(\textbf{MADA}) \cite{b13}, Conditional Domain Adversarial Network(\textbf{CDAN})\cite{b34}, Incremental CAN (\textbf{iCAN}) \cite{b38}, Domain Symmetric  Networks (\textbf{SymNets}) \cite{b32}, Generate To Adapt(\textbf{GTA}) \cite{b39} and Joint Domain alignment and Discriminative feature learning (\textbf{JDDA}) \cite{b33}.
\end{itemize}
% \vspace{-0.3cm}
\subsection{Experimental Setup}
% \vspace{-0.2cm}
To fairly compare our method with the state-of-the-art methods, we adopt the deep features commonly used in existing  unsupervised domain adaption methods. Specifically, DeCaf6 \cite{b59} features (activations of the 6th fully connected layer of a convolutional neural network trained on ImageNet, $d$ = 4096) are used for  Office-Caltech dataset, ResNet50 \cite{b3} features ($d$ = 2048) are used for Office-31 dataset and ImageCLEF-DA dataset. In this way, we can compare our proposed method with these deep models.

In our experiments, we adopt the PCA algorithm to decrease the dimension of the data before learning to accelerate the proposed method. We set the dimensionality of PCA space $m = 128$ for Office-Caltech dataset and $m = 256$ for Office-31 and ImageCLEF-DA datasets. For the dimensionality of the transformation matrix $P$, we set $k = 32, 128$ and $64$ for Office-Caltech, Office-31 and ImageCLEF-DA respectively. The number of iterations for CDEM to converge is $T = 11$ for all datasets. For regularization parameter $\delta$, we set $\delta = 1$ for Office-Caltech and ImageCLEF-DA datasets and $\delta = 0.1$ for Office-31 dataset. As for the other  hyper-parameters, we set $\beta, \lambda, \gamma$ and $\eta$ by searching through the grid with a range of $\{0.0001,0.001,0.01,0.1,1,10\}$. In addition, the coming experiment on parameter sensitivity shows that  our method can keep robustness with a wide range of parameter values.

\setlength{\abovecaptionskip}{-0.4cm}  % 调整图片标题与图距离
\setlength{\belowcaptionskip}{-0.5cm}   % 调整图片标题与下文距离
\begin{table*}[!t]
	\centering
	\caption{Classification Accuracy (\%) on Office-Caltech dataset using Decaf6 features.}
	\label{res_office_caltech}
	\resizebox{0.95\columnwidth}{!}{%
		\begin{tabular}{lccccccccccccc}
			\hline
			Method & C$\to$A & C$\to$W & C$\to$D & A$\to$C&A$\to$W & A$\to$D&W$\to$C & W$\to$A & W$\to$D & D$\to$C & D$\to$A & D$\to$W & Average \\ \hline
			DDC\cite{b29}  & 91.9 & 85.4 & 88.8 & 85.0 & 86.1 & 89.0 & 78.0 & 84.9 & \textbf{100.0} & 81.1 & 89.5 & 98.2 & 88.2\\
			DAN\cite{b9}  & 92.0 & 90.6 & 89.3 & 84.1 & 91.8 & 91.7 & 81.2 & 92.1 & \textbf{100.0} & 80.3 & 90.0 & 98.5 & 90.1\\
			DCORAL\cite{b25} & 92.4 & 91.1 & 91.4 & 84.7& - & - & 79.3 & - & - & 82.8 & - & - & -\\
			\hline
			%JDA\cite{long2013transfer}  & 89.6 & 85.1 & 89.8 & 83.6 & 78.3 & 80.3 & 84.8 & 90.3 & \textbf{100.0} & 85.5 & 91.7 & 99.7 & 88.2\\
			1NN \cite{Delany2007kNearestNC}& 87.3 & 72.5 & 79.6 & 71.7 & 68.1& 74.5& 55.3&62.6&98.1&42.1&50.0&91.5&71.1 \\
			SVM \cite{Evgeniou2001SupportVM} & 91.6 & 80.7 & 86.0 & 82.2 & 71.9& 80.9&67.9&73.4&\textbf{100.0}&72.8&78.7&98.3&82.0 \\
			PCA \cite{Hotelling1933AnalysisOA} & 88.1 & 83.4 & 84.1&79.3&70.9&82.2&70.3&73.5&\underline{99.4}&71.7&79.2&98.0&81.7\\ 
			TCA \cite{b8}& 89.8 & 78.3 & 85.4 & 82.6 & 74.2 & 81.5 & 80.4 & 84.1 & \textbf{100.0} & 82.3& 89.1& \underline{99.7} & 85.6\\
			JDA \cite{b10} & 89.6 & 85.1 & 89.8 & 83.6 & 78.3 & 80.3 & 84.8 & 90.3 & 100.0 & 85.5 & 91.7 & \underline{99.7} & 88.2 \\
			CORAL\cite{b54}& 92.0 & 80.0 & 84.7 & 83.2 & 74.6 & 84.1 & 75.5 & 81.2 & \textbf{100.0} & 76.8 & 85.5 & 99.3 & 84.7\\
			% SCA\cite{b24}  & 89.5 & 85.4 & 87.9 & 78.8 & 75.9 & 85.4 & 74.8 & 86.1 & \textbf{100.0} & 78.1 & 90.0 & 98.6 & 85.9 \\
			JGSA\cite{b26} & 91.4 & 86.8 & 93.6 & 84.9 & 81.0 & 88.5 & 85.0 & 90.7 & \textbf{100.0} & 86.2 & 92.0 & \underline{99.7} & 90.0\\
			MEDA\cite{b22} & \underline{93.4} & \underline{95.6} & 91.1 & \underline{87.4} & 88.1 & 88.1 & \textbf{93.2} & \textbf{99.4} & \underline{99.4} & 87.5 & \underline{93.2} & 97.6 & 92.8\\
			%SLPP & 91.3 & 73.6 & 86.6 & 82.6 & 72.2 & 82.8 & 71.8 & 79.5 & 100.0 & 79.2 & 88.5 & 99.3 & 84.0\\
			%CAPLS(LDA) & 91.1 & 85.4 & 94.9 & 83.5 & 86.4 & 90.4 & 87.7 & 92.5 & 100.0 & 87.8 & 92.4 & 99.7 & 91.0\\ \hline
			CAPLS \cite{b31} & 90.8 & 85.4 & 95.5 & 86.1 & 87.1 & \underline{94.9} & 88.2 & 92.3 & \textbf{100.0} & \underline{88.8} & 93.0 & \textbf{100.0}& 91.8\\
			%SPL-NCP (Ours) & 92.1 & 88.8 & 94.9 & 86.6 & 90.5 & 84.7 & 83.5 & 92.1 & \textbf{100.0} & 85.9 & 92.6 & 98.3 & 90.8\\
			%SPL-SP (Ours) & 92.4 & \underline{93.2} & \textbf{98.7} & \textbf{87.7} & \textbf{95.3} & 89.2 & 87.3 & 92.1 & \textbf{100.0} & 88.3 & 92.7 & 98.6 & \textbf{93.0}\\
			SPL\cite{b55}&92.7 & 93.2 & \textbf{98.7} &\underline{87.4} & \underline{95.3} & 89.2 & 87.0 & 92.0 & \textbf{100.0} & 88.6 & 92.9 & 98.6 & \underline{93.0}\\
			\hline
			CDEM (Ours) & \textbf{93.5} & \textbf{97.0} & \underline{96.2} & \textbf{88.7} & \textbf{98.0} & \textbf{95.5} & 
			\underline{89.1} & \underline{93.5} & \textbf{100.0} & \textbf{90.1} & \textbf{93.4} & \underline{99.7} & \textbf{94.6}\\
			\hline
		\end{tabular}%
	}
	\vspace{-0.5cm}
\end{table*}
% \vspace{-0.4cm}
\subsection{Results and Analysis}
% \vspace{-0.2cm}

The results on the \emph{Office-Caltech} dataset are reported in Table \ref{res_office_caltech}, where the highest accuracy of each cross-domain task is boldfaced. The results of baselines are directly reported from original papers if the protocol is the same. The CDEM method significantly outperforms all the baseline methods on most transfer tasks (7 out of 12) in this dataset. It is desirable that CDEM promotes the classification accuracies significantly on hard transfer tasks, e.g., \textbf{A$\to$D} and \textbf{A$\to$W}, where the source and target domains are substantially different \cite{Saenko2010AdaptingVC}. Note that CDEM performs better than SPL in most tasks, which only learns domain-invariant features across domains.

\begin{table}[!t]
	\centering	
	\caption{Accuracy (\%) on Office-31 dataset using either ResNet50 features or ResNet50 based deep models.
	}
\label{res_office31}
	\resizebox{0.55\columnwidth}{!}{%
		\begin{tabular}{lccccccc}
			\hline
			Method & \scriptsize{A$\to$W} & \scriptsize{D$\to$W} & \scriptsize{W$\to$D} & \scriptsize{A$\to$D} & \scriptsize{D$\to$A} & \scriptsize{W$\to$A} & Avg \\ \hline
			%DAN\cite{long2015learning}  & 80.5 & 97.1 & 99.6 & 78.6 & 63.6 & 62.8 & 80.4 \\
			RTN\cite{b40} & 84.5 & 96.8 & 99.4 & 77.5 & 66.2 & 64.8 & 81.6\\
			MADA\cite{b13} & 90.0 & 97.4 & 99.6 & 87.8 & 70.3 & 66.4 & 85.2 \\
			GTA \scriptsize{\cite{b39}} & 89.5& 97.9 & \underline{99.8}& 87.7 & 72.8 & 71.4& 86.5\\
			iCAN\cite{b38} & 92.5 & \textbf{98.8} & \textbf{100.0} & 90.1 & 72.1 & 69.9 & 87.2 \\
			CDAN-E\cite{b34} & \underline{94.1} & 98.6 & \textbf{100.0} & 92.9 & 71.0 & 69.3 & 87.7\\
			JDDA\cite{b33} & 82.6 & 95.2 & 99.7 & 79.8 & 57.4 & 66.7 & 80.2\\
			SymNets\cite{b32} & 90.8 & \textbf{98.8} & \textbf{100.0} & \underline{93.9} & 74.6 & 72.5 & \underline{88.4}\\
			TADA \cite{b11} &  \textbf{94.3} & \underline{98.7} & \underline{99.8} & 91.6 & 72.9 & 73.0 & \underline{88.4}\\
			\hline
			MEDA\cite{b22} & 86.2 & 97.2 & 99.4 & 85.3 & 72.4 & 74.0 & 85.7\\
			%SLPP & 77.9 &97.4 &99.2 &80.1 & 68.4 & 66.2 & 81.5\\
			%CAPLS(LDA)+minMeanDist & 76.9 & 99.1 & 99.8 & 79.5 & 61.9 & 60.5 & 79.6\\
			%CAPLS(Ours)+minMeanDist & 90.8 & 98.6 &  99.6  & 88.4 & \textbf{74.3} &\textbf{75.7} & \textbf{87.9} \\
			%CAPLS(LDA) & 77.0 & \textbf{99.1} & 99.8 & 77.9 & 61.8 & 60.8 & 79.4 \\ \hline
			CAPLS \scriptsize{\cite{b31}} & 90.6 & 98.6 & 99.6 & 88.6 & \underline{75.4} & \underline{76.3} & {88.2}\\
			%SPL-NCP (Ours) & 89.6 & \underline{98.7} & 99.6 & 89.4 & 74.3 & 75.5 & 87.8 \\
			%SPL-SP (Ours) & 93.0 & 98.6 & \underline{99.8} & \underline{93.0} & \textbf{76.4} &76.1 & \underline{89.5} \\
			% SPL \cite{b55}& 92.7 & \underline{98.7} & \underline{99.8} & 93.0 & \underline{76.4} & \underline{76.8} & \underline{89.6} \\ 
			\hline
			CDEM (Ours) & 91.1 & 98.4 & 99.2 & \textbf{94.0} & \textbf{77.1} & \textbf{79.4} & \textbf{89.9} \\
			\hline
		\end{tabular}%
		\vspace{-0.4cm}
	}
\end{table}
\begin{table}[!t]
	\centering
	\caption{Accuracy (\%) on ImageCLEF-DA dataset using either ResNet50 features or ResNet50 based deep models.}
	\label{res_image}	
	\resizebox{0.55\columnwidth}{!}{%		
		\begin{tabular}{lccccccc}
			\hline
			Method &I$\to$P & P$\to$I & I$\to$C & C$\to$I & C$\to$P & P$\to$C & Avg \\ \hline
			%DAN\cite{long2015learning}  & 75.0 & 86.2 & 93.3 & 84.1 & 69.8 & 91.3 & 83.3 \\
			%JDDA\cite{chen2018joint} & 82.6 & 95.2 & 99.7 & 79.8 & 57.4 & 66.7 & 80.2\\
			RTN\cite{b40} & 75.6 & 86.8 & 95.3 & 86.9 & 72.7 & 92.2 & 84.9\\
			MADA\cite{b13} & 75.0 & 87.9 & 96.0 & 88.8 & 75.2 & 92.2 & 85.8 \\
			%GTA\cite{sankaranarayanan2017generate} & 89.5& 97.9 & 99.8& 87.7 & 72.8 & 71.4& 86.5\\
			iCAN\cite{b38} & 79.5 & 89.7 & 94.7 & 89.9 & 78.5 & 92.0 & 87.4 \\
			CDAN-E\cite{b34} & 77.7 & 90.7 & \textbf{97.7} & 91.3 & 74.2 & 94.3 & 87.7\\
			SymNets\cite{b32} & \underline{80.2} & 93.6 & 97.0 & {93.4} & {78.7} & \underline{96.4} & 89.9\\
			\hline
			MEDA\cite{b22} & 79.7 & 92.5 & 95.7 & 92.2 & 78.5 & 95.5 & 89.0\\
			%SPL-NCP (Ours) & \underline{79.9} & 92.3 & 94.8 & 94.0 & \textbf{80.2} & 93.0 & 89.0 \\
			%SPL-SP (Ours) & 78.9 & \underline{94.3} & 96.8 & \underline{95.7} &\underline{78.9} & \textbf{96.8} & \underline{90.2} \\
			SPL \cite{b55}  & 78.3 & \underline{94.5} & 96.7 & \underline{95.7} & \underline{80.5} & 96.3 & \underline{90.3}\\
			\hline
			CDEM (ours)  & \textbf{80.5} & \textbf{96.0} & \underline{97.2} & \textbf{96.3} & \textbf{82.1} & \textbf{96.8} & \textbf{91.5}\\
			\hline	
		\end{tabular}%
	}	       
	\vspace{-0.6cm}                                            
\end{table}

The results on \emph{Office-31} dataset are reported in Table \ref{res_office31}. The CDEM method outperforms the comparison methods on most transfer tasks. Compared with the best shallow baseline method (CAPLS), the accuracy is improved by 1.7\%. Note that the CDEM method outperforms some deep domain adaptation methods, which implies the performance of CDEM in domain adaptation is better than several deep methods.

The results on \emph{ImageCLEF-DA} dataset are reported in Table \ref{res_image}. The CDEM method substantially outperforms the comparison methods on most transfer tasks, and with more rooms for improvement. An interpretation is that the three domains in \emph{ImageCLEF-DA} are visually dissimilar with each other, and are difficult in each domain with much lower in-domain classification accuracy \cite{b10}. MEDA and SPL are the representative shallow domain adaptation methods, which both focus on learning domain-invariant features. Moreover, SPL also uses selective target samples for adaptation. Consequently, the better performance of CDEM implies that minimizing cross-domain errors can further reduce the discrepancy across domains and achieve better adaptation.
% \vspace{-0.5cm}
\subsection{Effectiveness Analysis}
% \vspace{-0.3cm}
\subsubsection{Ablation Study} 
 We conduct an ablation study to analyse how different components of our method contribute to the final performance. When learning the final classifier, CDEM involves four components: the empirical error minimization (ERM), the distribution alignment (DA), the cross-domain error minimization (CDE) and discriminative feature learning (DFL) . We empirically evaluate the importance of each component. To this end, we investigate different combinations of four components and  report average classification accuracy on three datasets in Table \ref{res_ab}. Note that the result of the first setting (only ERM used) is like the result of the source-only method, where no adaptation is performed across domains. It can be observed that methods with distribution alignment or cross-domain error minimization outperform those without distribution alignment or cross-domain error minimization. Moreover, discriminative learning can further improve performance and CDE achieves the biggest improvement compared with other components. Summarily, using all the terms together achieves the best performance in all tasks.
\setlength{\abovecaptionskip}{-0.9cm}  % 调整图片标题与图距离
\setlength{\belowcaptionskip}{0.2cm}   % 调整图片标题与下文距离
\begin{table}[tbp]
	\centering
	{%\large
		\centering
		\caption{Results of ablation study.}
		\label{res_ab}
		\resizebox{0.55\columnwidth}{!}{%
			\begin{tabular}{c|c|c|c|cccc}
				\toprule
				% \multicolumn{4}{c|}{Method} & Office-Caltech & Office31 & ImageCLEF-DA  \\ 
				% \hline
				% % \multicolumn{4}{c|}{Source only} & & & & \\
				% \hline
				%  ERM & DA & CDE & DFL & & & & \\
				\multicolumn{4}{c|}{Method} & \multirow{2}{*}{\footnotesize{Office-Caltech}} & \multirow{2}{*}{\footnotesize{Office31}} & \multirow{2}{*}{\footnotesize{ImageCLEF-DA}}  \\ 
				\cline{1-4} ERM & DA & CDE & DFL & & & & \\ \hline
				\cmark & \xmark & \xmark & \xmark  & 90.2 & 86.6  & 87.5  \\ 
				\cmark & \cmark & \xmark & \xmark & 91.5 & 87.2 & 88.6  \\
				% \cmark & \xmark & \cmark & \xmark& 93.7  & 88.7 & 90.6 \\
				% \cmark & \xmark & \xmark & \cmark & 93.5 & 88.8 & 90.5 \\ \hline
				\cmark & \cmark & \cmark & \xmark& 94.0 & 89.2 & 90.8 \\
				\cmark & \cmark & \cmark & \cmark & 94.6 & 89.9 & 91.5 \\
				\bottomrule
			\end{tabular}%
		}
	}
	\vspace{-0.8cm}
\end{table}
\vspace{-0.6cm}
\subsubsection{Evaluation of Selective Target Samples}
We further perform experiments to show the effectiveness of selective target samples. We compare several variants of the proposed method: \textbf{a}) No selection: We use all the target samples for training without any samples removed. \textbf{b}) Only label consistency: We only select the samples where the predicted label by $p_s(x_t)$ is the same with $p_t(x_t)$. \textbf{c}) Only high probabilities: We only select the target samples with high prediction confidence. \textbf{d}) The proposed method. As shown in Fig \ref{acc_param}(a), “No selection” leads to a model with the worst performance due to the catastrophic error accumulation. The "Only label consistency" and "Only high probabilities" achieve significantly better results than “No selection”, but are still worse than the proposed method, which verifies that our method of explicitly selecting easier samples can make the model more adaptive and less likely to be affected by the incorrect pseudo labels.

\vspace{-0.4cm}
\subsubsection{Feature Visualization.} 
\begin{figure*}[tbp]
	 \centering
	 \subfigure[\scriptsize{Ablation study of selective samples}]{
	 \includegraphics[width = 0.22\textwidth]{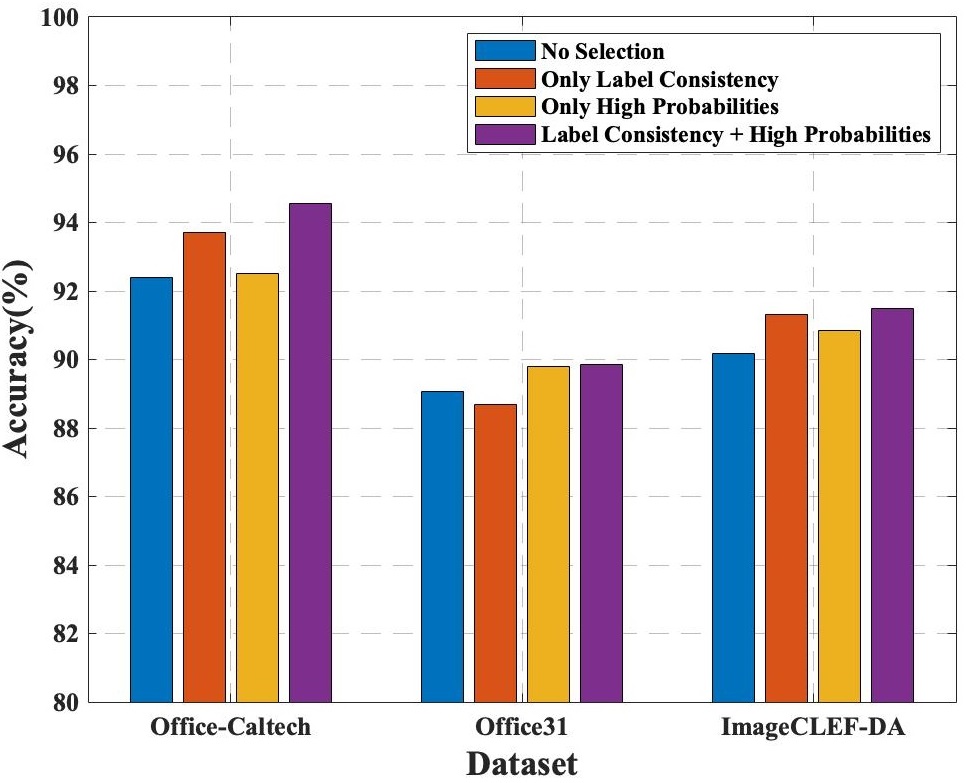}
	 }
	 \subfigure[t-SNE before adaptation]{
	 \includegraphics[width = 0.22\textwidth]{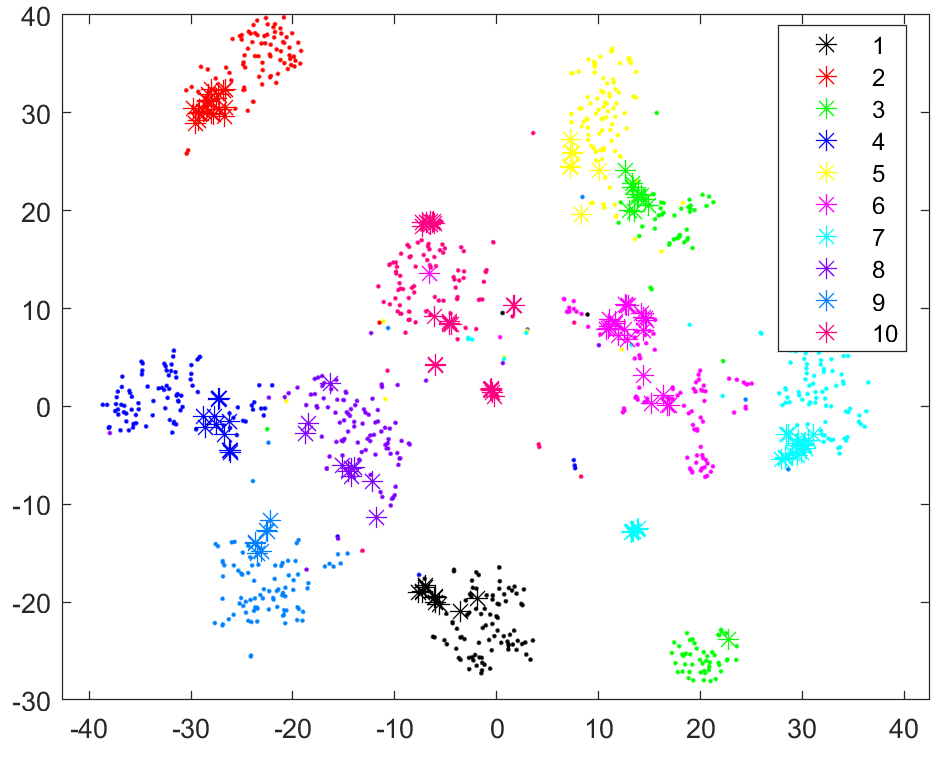}
	 }
	 \subfigure[t-SNE by JDA]{
	 \includegraphics[width = 0.225\textwidth]{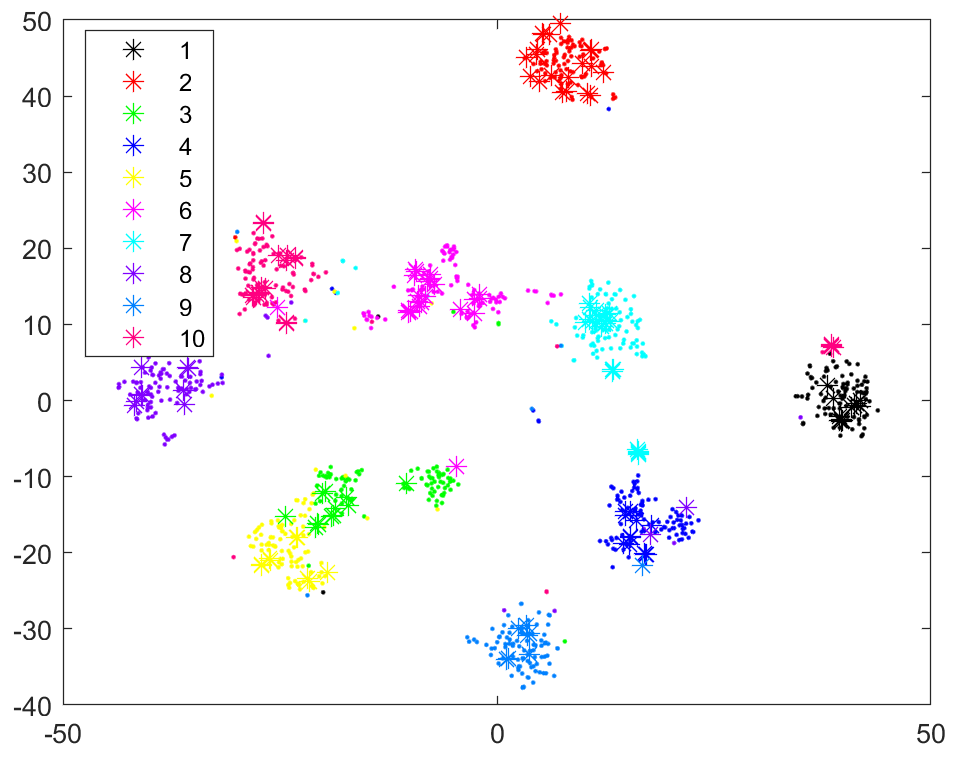}
	 }
	 \subfigure[t-SNE by CDEM]{
		 \includegraphics[width = 0.22\textwidth]{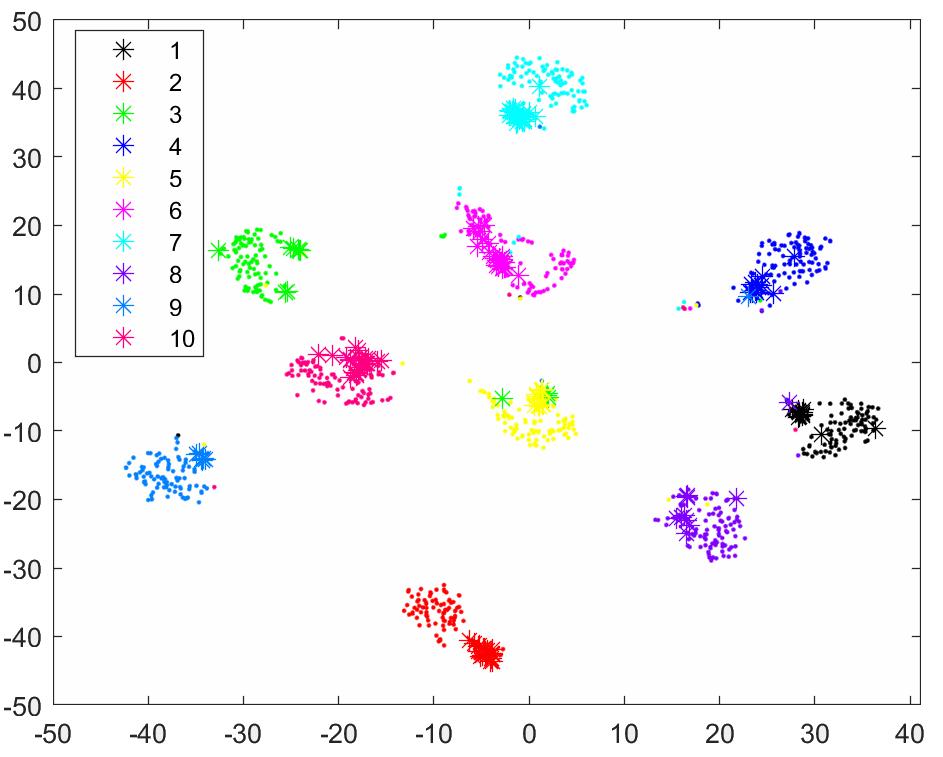}
	 }
	%  \subfigure[m]{
	%  \includegraphics[width = 0.215\textwidth]{figures/5.png}
	%  }
	%  \subfigure[k]{
	%  \includegraphics[width = 0.23\textwidth]{figures/1.png}
	%  }
	%  \subfigure[T]{
	%  \includegraphics[width = 0.225\textwidth]{figures/2.png}
	%  }
	%  \subfigure[$\gamma$]{
	%  \includegraphics[width = 0.215\textwidth]{figures/4.png}
	%  }
 \caption{ Ablation study of selective samples, t-SNE visualization and parameter sensitivity}
 \label{acc_param}
 \vspace{-0.4cm}
 \end{figure*}

In Fig \ref{acc_param}(b-d), we visualize the feature representations of task \textbf{A}$\rightarrow$\textbf{D} (10 classes) by t-SNE \cite{sne} as previous methods \cite{b55} using JDA and CDEM. Before adaptation, we can see that there is a large discrepancy across domains. After adaptation, JDA learns domain-invariant features which can reduce distribution discrepancy, the source domain and the target domain can become closer. While  CDEM further considers the cross-domain errors, achieving a better performance.

% \vspace{-0.35cm}
% \subsection{Parameter Sensitivity}
%  In this subsection, we only evaluate CDEM with a wide range of values for the cross-domain error parameter $\gamma$, the dimension of PCA space $m$, the dimension of transformation matrix $k$ and the number of iterations $T$ due to space limitation.  The results are shown in Fig \ref{acc_param}(e)-(h). As we can see, CDEM can achieve a robust performance concerning a wide range of parameter values.  Specifically, smaller cross-domain error parameters and a proper dimension of both PCA space $m$ and CDEM space $k$ in the range of $[64,128]$ can obtain better performance.

	 % \subfigure[$lambda$]{
	 % \includegraphics[width = 0.22\textwidth]{3.png}
	 % }
\vspace{-0.4cm}	 
\section{Conclusion}
\vspace{-0.2cm}
In this paper, we propose the \emph{Cross-Domain Error Minimization} (CDEM), which not only learns domain-invariant features across domains but also performs cross-domain error minimization. These two goals complement each other and contribute to better domain adaptation. Apart from these two goals, we also integrate the empirical error minimization and discriminative learning into a unified learning process. Moreover, we propose a method to select the target samples to alleviate error accumulation problem caused by incorrect pseudo labels. Through a large number of experiments, it is proved that our method is superior to other strong baseline methods.

\section{Acknowledgements}
This paper is supported by the National Key Research and Development Program of China (Grant No. 2018YFB1403400), the National Natural Science Foundation of China (Grant No. 61876080), the Collaborative Innovation Center of Novel Software Technology and Industrialization at Nanjing University.

\setlength{\abovecaptionskip}{0.0cm}  % 调整图片标题与图距离
 \setlength{\belowcaptionskip}{-0.2cm}   % 调整图片标题与下文距离
\vspace{-0.3cm}
\bibliographystyle{splncs04}
\bibliography{DASFAA_1}

\end{document}